\documentclass[sigconf, screen, nonacm]{acmart}
\usepackage[utf8]{inputenc}
\usepackage{cleveref}
\usepackage{bbm}
\usepackage{pifont}
\usepackage{multirow}
\crefname{section}{§}{§§}
\Crefname{section}{§}{§§}
\renewcommand\footnotetextcopyrightpermission[1]{}

\AtBeginDocument{%
  }

\setcopyright{acmlicensed}
\copyrightyear{2025}
\acmYear{2025}
\acmDOI{XXXXXXX.XXXXXXX}
\acmConference[MM '25]{Proceedings of the 33th ACM International Conference on Multimedia}{October 27--31, 2025}{Dublin, Ireland}
\acmISBN{978-1-4503-XXXX-X/2025/06}

\acmSubmissionID{xxxx}

\begin{document}

\title{FocusedAD: Character-centric Movie Audio Description}

\author{Xiaojun Ye}
\authornote{Both authors contributed equally to this research.}
\orcid{0009-0008-2490-8247}
\affiliation{%
  \institution{Zhejiang University}
  \country{China}
}
\email{yexiaojun@zju.edu.cn}

\author{Chun Wang}
\authornotemark[1]
\orcid{0009-0003-7812-0520}
\affiliation{%
  \institution{Zhejiang University}
  \country{China}
}
\email{zjuheadmaster@zju.edu.cn}

\author{Yiren Song}
\affiliation{%
  \institution{National University of Singapore}
  \country{Singapore}
}
\email{yiren@nus.edu.sg}

\author{Sheng Zhou}
\authornote{Corresponding author.}
\affiliation{%
  \institution{Zhejiang University}
  \country{China}}
\email{zhousheng_zju@zju.edu.cn}

\author{Liangcheng Li}
\affiliation{%
  \institution{Zhejiang University}
  \country{China}}
\email{liangcheng_li@zju.edu.cn}

\author{Jiajun Bu}
\affiliation{%
  \institution{Zhejiang University}
  \country{China}}
\email{bjj@zju.edu.cn}

\renewcommand{\shortauthors}{Xiaojun Ye, Chun Wang et al.}

\begin{abstract}
Movie Audio Description (AD) aims to narrate visual content during dialogue-free segments, particularly benefiting blind and visually impaired (BVI) audiences. Compared with general video captioning, AD demands plot-relevant narration with explicit character name references, posing unique challenges in movie understanding.To identify active main characters and focus on storyline-relevant regions, we propose FocusedAD, a novel framework that delivers character-centric movie audio descriptions. It includes: (i) a Character Perception Module(CPM) for tracking character regions and linking them to names; (ii) a Dynamic Prior Module(DPM) that injects contextual cues from prior ADs and subtitles via learnable soft prompts; and (iii) a Focused Caption Module(FCM) that generates narrations enriched with plot-relevant details and named characters. To overcome limitations in character identification, we also introduce an automated pipeline for building character query banks. FocusedAD achieves state-of-the-art performance on multiple benchmarks, including strong zero-shot results on MAD-eval-Named and our newly proposed Cinepile-AD dataset. Code and data will be released at \url{https://github.com/Thorin215/FocusedAD}.
\end{abstract}




\maketitle

\section{Introduction}
\begin{figure}[h]
  \centering
  \includegraphics[width=\linewidth]{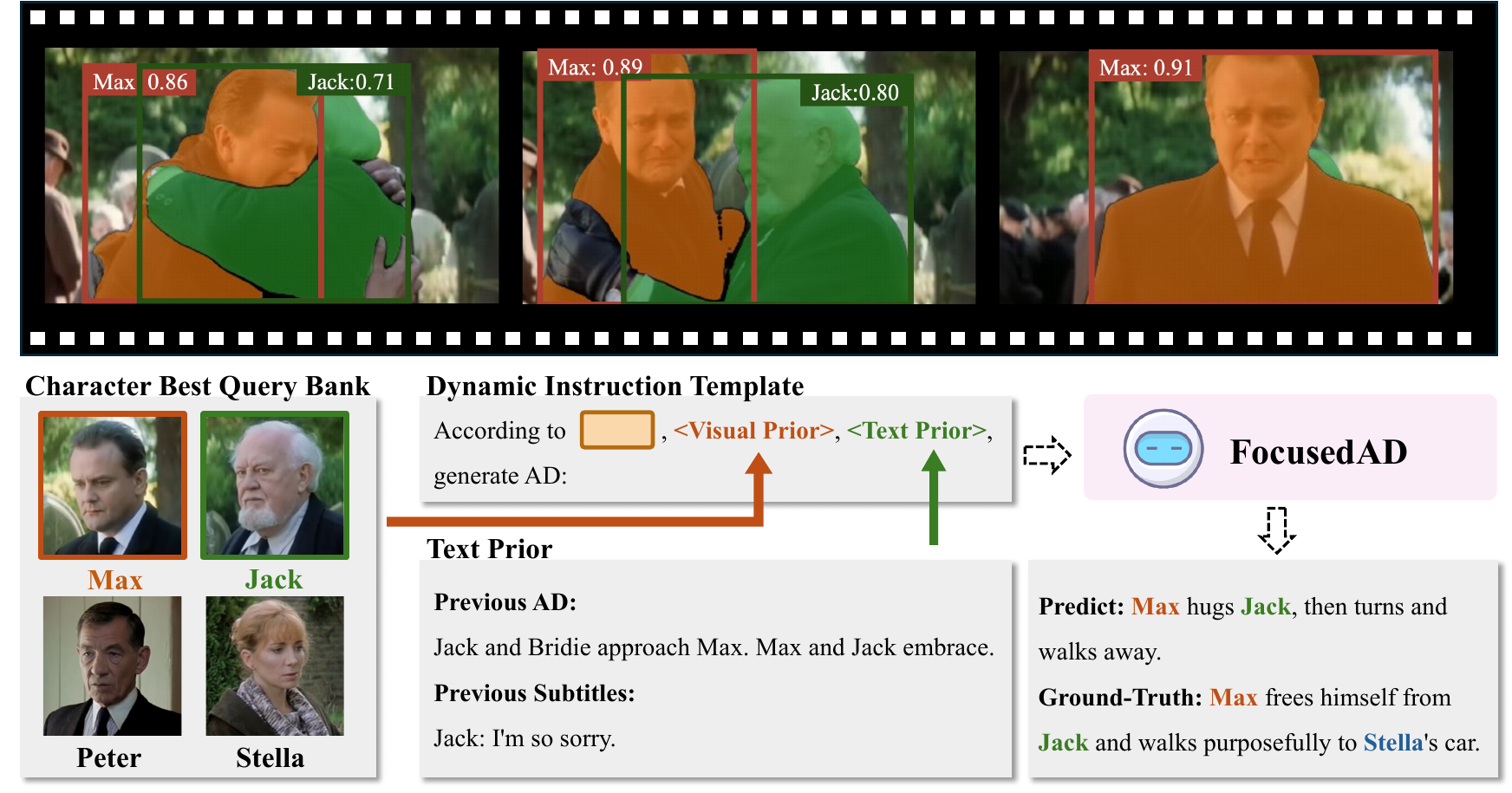}
  \caption{FocusedAD: We propose an automated character-centric AD generation model that emphasizes main character regions' appearances and actions while incorporating narrative context. Characters appearing in the movie clip are annotated with colored bounding boxes.}
  \label{fig:intro}
\end{figure}

Movie Audio Description (AD) is a narration service that verbally describes visual content in video content that are not conveyed through the existing soundtrack~\cite{han2023autoad}. 
AD is developed to primarily assist Blind and Visually Impaired (BVI) audiences in understanding movie storylines~\cite{perego2016gains}. 
Due to the prohibitive cost of employing skilled annotators for AD creation, the amount of AD-equipped videos remains limited, highlighting the need for automatic AD generation systems. 
Recent advancements in multi-modal large language models (MLLMs) for video understanding~\cite{zhou2024streaming,chen2024sharegpt4video} have made automatic movie audio description generation feasible.

BVI audiences need to simultaneously process both AD and the original soundtrack including dialogues and ambient sounds, resulting in high cognitive load. Therefore, AD narratives must be concise to prevent information overload.
Due to the length constraints, AD needs to compress complex visual information into limited textual descriptions while preserving narrative crucial elements~\cite{rohrbach2017movie} such as characters' description~\cite{kukleva2020learning} and scene transitions~\cite{ACB}.
Particularly in character-centric scenes, the appearance, actions, and expressions of individual characters, along with inter-character interactions and relationships, carry greater narrative significance than static scene arrangements and plot-irrelevant objects~\cite{yu2021transitional}.
However, existing works~\cite{han2023autoad,han2023autoadii,zhang2024mm} typically employ image encoders to extract key frame features followed by LLM  for generating dense video captions, resulting in relatively smooth attention distribution across scene-level features~\cite{yuan2024videorefer}. This methodology struggles to achieve an optimal balance between descriptive detail and narrative coherence, often producing redundant object-centric (e.g., "red chair, blue table") rather than event-driven storytelling (e.g., "the character anxiously surveys the room"). 
Thus, it is essential for automated AD generation models to possess the ability to focus on \textit{narrative-salient regions}, a feature that current methods inadequately address.

Furthermore, in order to facilitate the comprehension of the story for audiences, particularly individuals in the BVI, AD requires explicit character identification by name (e.g. "Harry walks in")~\cite{han2023autoadii}. 
This poses more stringent requirements on content description compared to conventional video captioning~\cite{seo2022end}, which employs ambiguous expressions such as "he/she" or "a man/woman".
Following character region detection, concurrent with description generation, the model ought to establish precise mappings between detected regions and character identities, thereby enabling effective character name references in the generated AD~\cite{zhang2024mm}.
This demands enhanced capabilities in character tracking and identification across different frames, particularly challenging in multi-character scenarios. 
However, certain models~\cite{han2023autoad,zhang2024mm} rely on subtitles to extract character names, where the frequency of character name occurrences is substantially lower than AD requirements. Other models~\cite{han2023autoadii,ye2024mmad} incorporate additional character recognition modules with inadequate performance due to feature shifts caused by variations in characters' appearances and makeup. These approaches struggle to effectively \textit{identify characters associated with ongoing actions}, resulting in semantic ambiguity and reduced narrative coherence.


In this paper, we present \textbf{FocusedAD}, a novel model that recognize effective main character and generate ad which focus on narratively relevant elements, as illustrated in ~\cref{fig:intro}. 
To enable the generated AD to focus on narrative-salient regions, we first introduce a \textbf{Character Perception Module (CPM)} to identify main characters in arbitrary frames of current movie clips. The module employs bi-directional propagation to obtain character-specific region lists throughout the movie clips. The detected character regions are subsequently fed into two distinct modules.
In order to address the challenges of active character recognition, we introduce an automated clustering-based pipeline for optimal character bank query generation, applicable to both our training and testing sets. With the character best query bank, we effectively identify main characters by computing distances between detected faces and each character's best query, incorporating this information as visual priors into the \textbf{Dynamic Prior Module (DPM)}, which dynamically adapts to scenes with varying numbers of characters.
Finally, we introduce the \textbf{Focused Caption Module (FCM)}, performing joint reasoning over scene visual tokens $\mathcal{T}_S$, character tokens $\mathcal{T}_C$, and text tokens $\mathcal{T}_W$ to generate character-centric ADs, which is then evaluated against ground-truth ADs. 
Our approach addresses a critical limitation of current MLLMs that tend to generate redundant content when processing video captioning tasks, resulting in movie ADs that more effectively facilitate story comprehension for BVI audiences.

Our primary contributions can be summarized as follows:

\begin{itemize}
    \item We propose FocusedAD, a model that leverages both temporal context and character information to generate story-coherent and character-centric AD. Through the soft prompt mechanism, our model dynamically adapts to scenes containing varying numbers of characters.
    \item We develop an automated data construction pipeline for creating an optimal character query bank, effectively addressing the limitation of insufficient character information in both training and testing AD datasets.
    \item We show promising results on character-centric AD, as seen from both qualitative and quantitative evaluations, and also achieve impressive zero-shot state-of-the-art performance on both the MAD-eval-Named benchmark and our newly introduced Cinepile-AD test set.
\end{itemize}

\begin{figure*}[h]
  \centering
  \includegraphics[width=\linewidth]{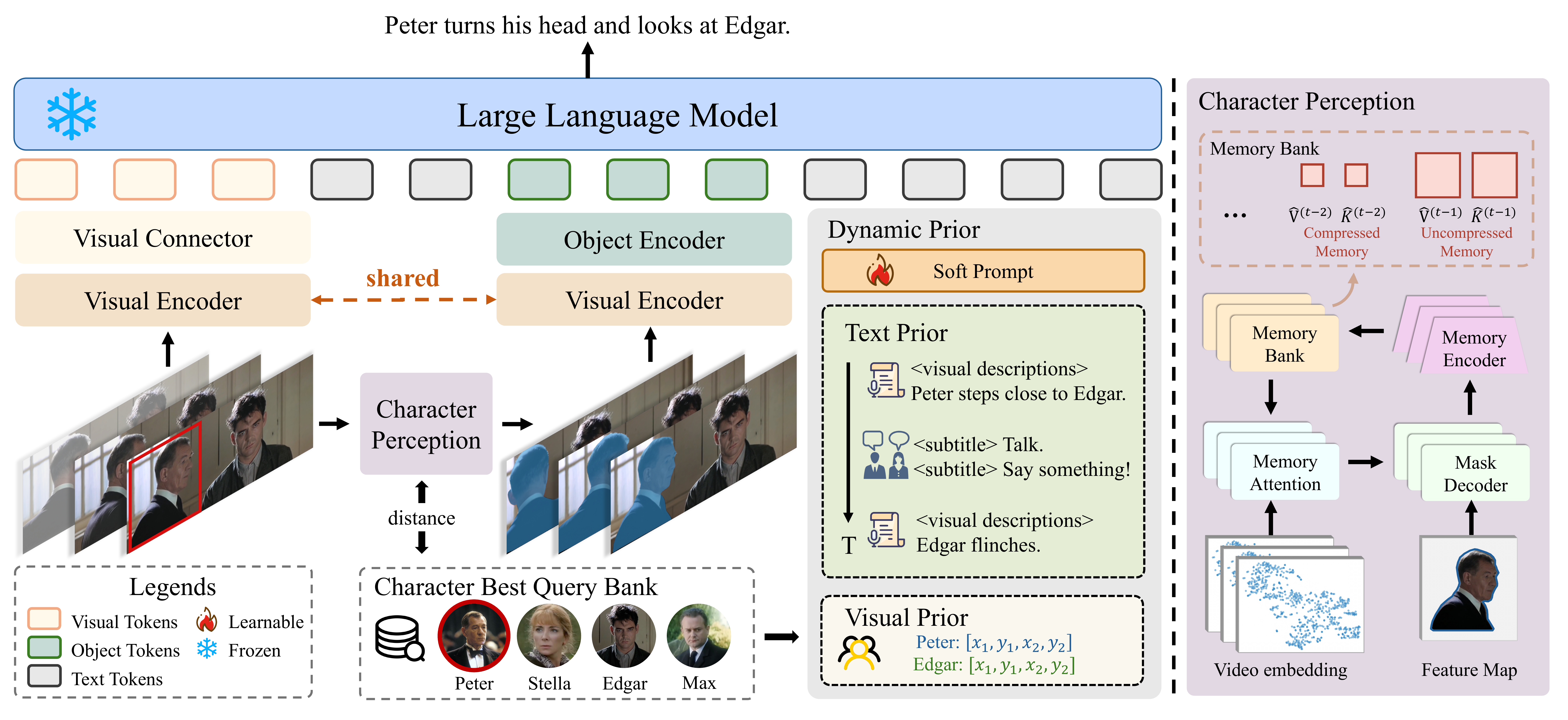}
  \caption{Overview of FocusedAD: FocusedAD takes movie clips as input and captures the character best query bank through clustering. The Character Perception Module identifies main characters in key frames and bi-directionally propagates character regions across the entire key frame sequence. Then, through the Dynamic Prior Module, it dynamically integrates visual and text priors using soft prompts. Finally, the Focused Caption Module takes scene-level tokens, character-level tokens, and soft prompts as input to generate character-centric audio descriptions.}
  \label{fig:architecture}
\end{figure*}

\section{Related Work}
\subsection{Video Caption}
Video captioning is the task of automatically generating natural language descriptions that accurately describe the visual content and temporal dynamics of video sequences~\cite{lin2022swinbert,luo2020univl}. A closely related task to AD is dense video captioning~\cite{krishna2017dense,mun2019streamlined,yang2023vid2seq}, which aims to generate multiple captions along with their corresponding temporal localization within the video. Recent methodologies~\cite{li2018jointly,chen2021towards,chadha2020iperceive} address this task through the joint optimization of captioning and localization modules, facilitating enhanced inter-task interactions.
However, dense video captioning typically describes discrete event segments in isolation, lacking the narrative coherence across segments, which is essential for generating AD. 

With the emergence of multi-modal Large Language Models (MLLMs)~\cite{alayrac2022flamingo,liu2023llava}, video caption has seen new approaches either through visual instruction tuning~\cite{li2023videochat} or LLM prompting~\cite{lin2023mm}.  Although MLLMs demonstrate more robust adaptability across diverse scenarios and can address domain discrepancies through Supervised Fine-tuning (SFT), existing models still rely heavily on global scene level features, resulting in their inability to balance between fine-grained descriptions and semantically irrelevant redundant narratives. 
Different from previous work, we propose to leverage Character Perception Module and Focused Caption Module to generate character-centric caption.

\subsection{Audio Description}
AD offers the narration describing visual elements in movies, expected to weave a coherent narrative for storyline~\cite{zhang2024mm}. Early research in the AD field focused primarily on data collection and preprocessing~\cite{torabi2015using,soldan2022mad,rohrbach2017movie}. 
However, existing AD-annotated movie datasets suffer from various limitations such as restricted keyframe-only content~\cite{huang2020movienet} or frame-level feature constraints~\cite{soldan2022mad}.
Initial studies of models concentrated on predicting AD insertion timestamps based on audiovisual content~\cite{wang2021toward} or scripts~\cite{soldan2022mad,campos2023machine}. Recent studies~\cite{han2023autoad,han2023autoadii,zhang2024mm} have explored the integration of visual features with LLMs for the description of visual information. However, these automated AD generation pipelines still rely on global features, causing models to distribute attention across the entire scene, rather than focus on the details relevant to the story.
Unlike previous works, our FocusedAD specifically focuses on narrative-salient regions and generates storyline-coherent descriptions.

Compared with general video captioning, AD is expected to incorporate effective character names into descriptions~\cite{brown2021automated,huang2018person}. Thus, character recognition serves as a prerequisite for the task, with numerous works proposing automatic identification pipelines based on face recognition~\cite{brown2021face} or ReID~\cite{xiao2017joint,li2023clip}. 
Current models are exploring methods to incorporate character names in generated text, either through Named Entity Recognition (NER) from subtitles~\cite{han2023autoad} or face recognition based on actor portraits~\cite{han2023autoadii}. 
Departing from previous approaches, we propose a method based on automated clustering to obtain a character best query bank for more robust character identity features generation, effectively addressing the variations in character visual features caused by changes in lighting conditions and scene contexts.

\section{Method}
Given a long-form movie $\mathcal{V}$ , which is segmented into multiple non-dialogue visual segments$\{x_1,x_2,...,x_T\}$ based on character dialogues, our research aims to automatically generate movie audio description (AD) in text form for these visual intervals. Each movie clip is cut from the raw movie based on the timestamp $[t_{start}, t_{end}]$ given by the AD annotation. Specifically, for the $i$-th movie clip consisting of multiple key frames $x_i=\{\mathcal{I}_1,\mathcal{I}_2,...,\mathcal{I}_K\}$, we aim to produce AD $T_i$ that narrates the visual events in such a way that helps the visually impaired follow the storyline. To this end, we present FocusedAD to generate character-centric AD. 
In the following sections, we first introduce the \textit {Character Perception Module}(~\cref{sec3.1}) (CPM), which performs facial grounding to locate main characters in arbitrary key frames, establishing associations between these regions and character names. The character names subsequently serve as visual priors for downstream tasks.
Next, we present the \textit{Dynamic Prior Module}(~\cref{sec3.2}) (DPM), which employs soft prompts to establish connections between visual priors generated by CPM and textual priors derived from previous AD $T_{t\textless i}$ and subtitles $S_{t\leq i}$.
These components feed into \textit{Focused Caption Module}(~\cref{sec3.3}) (FCM) that generates AD ${T_i}$, which focus on character-centric details.  The architecture of our model is shown in ~\cref{fig:architecture}.

\subsection{Character Perception Module}
\label{sec3.1}
\begin{figure}[h]
  \centering
  \includegraphics[width=\linewidth]{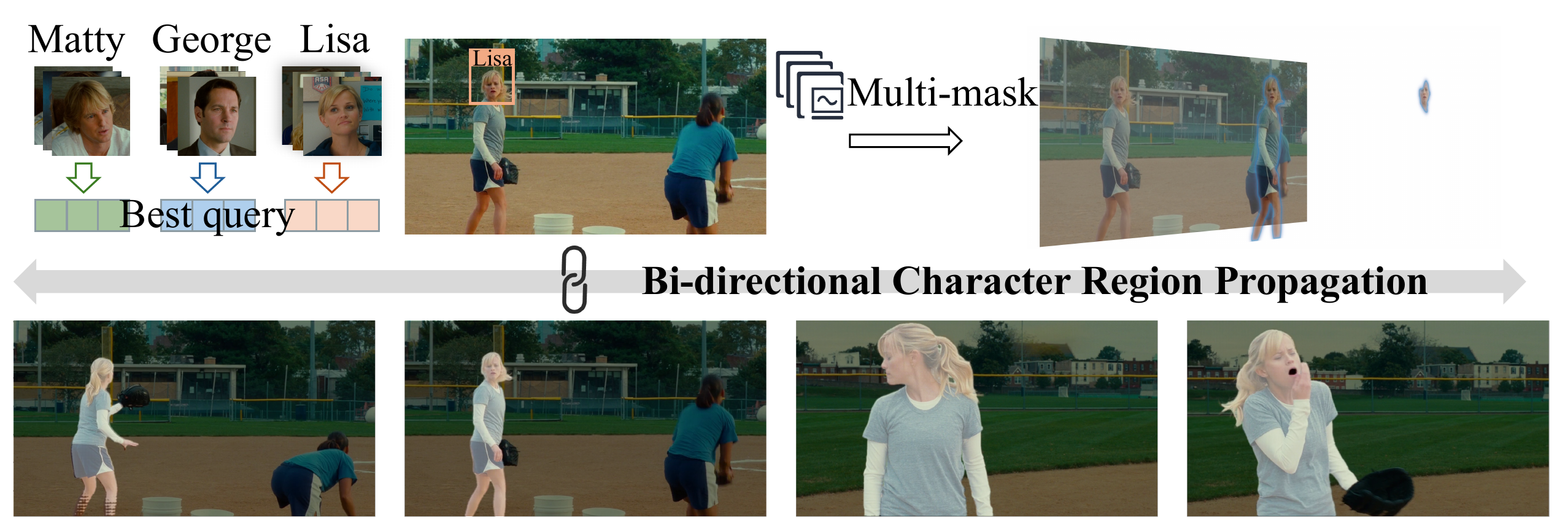}
  \caption{Character Perception Module traverses the key frame sequence, detecting main characters in any frame and obtaining their segmented regions. Videos are processed in a streaming fashion, where each frame cross-attends to the main character memories from context frames. Finally, both the region prediction and key frame embeddings are stored into memory bank.}
  \label{fig:character}
\end{figure}

As shown in ~\cref{fig:character}, given that narrative progression in audio descriptions derives from either scene transitions or main character activities, we implement a dual-strategy approach. For scenes without characters, the model describes environmental changes, while for movie clips containing characters, our goal is to identify active main characters in any frame of the input clips and output the frame indices where main characters are detected, guiding the model to focus on character actions and appearances. To this end, we (i) calculate character best queries by extracting character portraits from the movie, further calibrated by clustering algorithms, (ii) train a character detection module that predicts the active characters given their best queries and the current movie clip ,and (iii) propagate the detected main character region in both directions in multiple key frames.

\textbf{Character Query Clustering.}
Actors' portrait images can differ considerably in appearance from the character in the movie due to various factors, such as hairstyle, makeup, dress, ageing, or camera viewpoint ~\cite{nagrani2018benedict}. To improve the accuracy of character recognition, we propose a calibration mechanism to get the best character query. Initially, each movie contains $J$ main characters, For the $j$-th main character, we can build a portrait collection $P_j=\{p^j_1,p^j_2,...,p^j_n\}$, where $n$ denotes the number of cropped portraits. These portraits are encoded into visual feature embedding $E_j=\{e^j_1,e^j_2,...,e^j_n\}$. Our goal is to obtain an optimal set of character representations $Q=\{q_1,q_2,...,q_J\}$, where $q_j$ denotes the best query for the $j$-th main character.

Our approach begins by quantifying the visual coherence within each character identity through intra-class distance measurement:
\begin{equation}
D_\mathrm{int}(q_j)=\frac{1}{|E_j|}\sum_{e_i\in E_j}\|e_i-q_j\|_2
\end{equation}

Subsequently, we evaluate character distinctiveness by calculating the inter-class distance:
\begin{equation}
D_\mathrm{ext}(q_j)=\min_{j'\neq j}\|q_j-q_{j'}\|_2,
\end{equation}
where $q_{j'}$ denotes the best query for the $j'$-th character.

Finally, We compute the centroid of each cluster as the best query list $Q$ by maximizing the following objective function:
\begin{equation}
f(Q)=\sum_{j=1}^m\left(\frac{D_{\mathrm{ext}}(q_j)}{D_{\mathrm{int}}(q_j)+\epsilon}\right),
\end{equation}
where $\epsilon=10^{-6}$ to prevent division by zero.

\textbf{Character Recognition.}
Given a sequence of key frames $=\{\mathcal{I}_1,\mathcal{I}_2,...,\mathcal{I}_K\}$ for the movie clip $x_i$, we perform main character recognition by iterating through each key frame. Specially, we process the video frame by frame using MTCNN~\cite{zhang2016joint} as our face detector to locate facial regions in each frame. For frame $k$ , assuming $C$ faces are detected, we obtain a set of cropped face images $\{x^k_1,x^k_2,...,x^k_A\}$. Each detected face image is mapped to a 128-dimensional Euclidean space through the FaceNet~\cite{schroff2015facenet}:
\begin{equation}
f(x^k_a)\in\mathbb{R}^{128},\quad\mathrm{s.t.~}\|f(x^k_a)\|_2=1,
\end{equation}
where $x^k_a$ denotes the $a$-th face (anchor) detected in $x^k$-th key frame.
The model is optimized using triplet loss to ensure embeddings from the same identity are close while different identities are separated in the embedding space.
For each detected face embedding $f(x^k_a)$, we compute the squared $L2$ distance to each character best query:
\begin{equation}
D(f(x^k_a),q_j)=\|f(x^k_a)-q_j\|_2^2, q_j \in Q
\end{equation}

We employ a threshold $u$ to determine the corresponding character name $C^k_a$ in the main character list for the current face $y^k_a$:

\begin{equation}
C_a^k=
\begin{cases}
\underset{j\in {0,\ldots,J-1}} {\arg\min} D(f(y_a^k),q_j), & \mathrm{if~}\underset{j\in {0,\ldots,J-1}}\min D(f(y_a^k),q_j)<u \\
\mathrm{Unknown}, & \mathrm{otherwise} 
\end{cases}
\end{equation}

\textbf{Temporal Character Region Propagation.}
Our Temporal Character Region Propagation Module allows the character regions on any frame of the movie clip as input. After receiving initial region, the module propagates these regions to obtain the masklet of the character across the entire movie clip, localizing the segmentation mask of the target on every key frame. Specially, we firstly share the visual features extracted from an MAE~\cite{he2022masked} pre-trained Hiera~\cite{ryali2023hiera} image encoder across all objects and global features within the movie clip. Then, we utilize the memory encoder~\cite{ravi2024sam} to generate a memory by downsampling the output mask using a convolutional module and summing it element-wise with the unconditioned frame embedding from the image encoder, followed by light-weight convolutional layers to fuse the information.
Meanwhile, we maintain a memory bank that keeps track of past predictions for the target character in the movie clip by maintaining a FIFO queue of memories from up to $K'$ context frames, and stores prompt information in a FIFO queue of up to the frames which recognizes main characters.
The memory attention operation takes the per-frame embedding from the image encoder and conditions it on the memory bank, before the mask decoder ingests it to form a prediction.
Finally, we employ memory attention to condition the current frame features on both historical frame features and predictions, as well as any newly detected character regions. The architecture consists of $L$ stacked transformer blocks, with the first block taking the image encoding of context frames as input. Each block sequentially performs self-attention, followed by cross-attention to frame memories and character regions stored in the memory bank, and concludes with an MLP layer.

\subsection{Dynamic Prior Module}
\label{sec3.2}
Compared to video caption where the annotation describes "what is in the video", movie AD describes the visual events in the scene that are relevant to the broader story, often centered around events, characters and the interactions between them.  To tackle these temporal dependencies, we propose to include two components to incorporate the essential contextual information from movies: (i)previous AD and subtitles, (ii)dynamic template prompt for both single and multi main character scene.

\textbf{Text Prior.} 
Events occurring in the current movie clips are not temporally independent. Incorporating previous audio descriptions and subtitles as background context helps the model maintain narrative continuity.  Specially, Our model takes the previous ADs $T_{t\textless i}$ and intervening subtitles $S_{t\leq i}$ to generate the AD $T_i$ for the current clip. The past movie ADs and subtitles are a few sentences, which are first concatenated into a single paragraph, then tokenized and converted to a sequence of word embeddings $W$. 

\textbf{Dynamic instruction template.}
Given the number $A$ of detected character regions in the $k$ key frame and the corresponding set of character names $\{C_1,C_2,...,C_{A'}\}, A'\leq A$, where $A'$ represents the number of main characters among the $A$ detected faces. We define the dynamic prompt generation function $f_{Dit}$ as:

\begin{equation}
f_{Dit}(A',\{C_i\})=
\begin{cases}
B & \text{if }A'=0 \\
C_1+B & \text{if }A'=1 \\
M+\sum_{i=1}^{A'}C_i+B & \text{if }A'>1
\end{cases},
\end{equation}
where $B$ denotes the base description prompt, $C_i$ denotes the prompt template for the i-th character, and $M$ denotes the prefix prompt for multi-character scenes.

\textbf{Soft prompt.} While the visual encoders in MLLMs have been trained on large-scale video-text pairs, they may not be sufficient to comprehensively extract all crucial visual information required for the complex task of movie audio description. It has been demonstrated that increasing the amount and quality of pretraining multi-modal datasets can significantly improve the movie story understanding capability of MLLMs. An attractive alternative approach is to leverage existing AD datasets for descriptive style fine-tuning.

\begin{figure}[h]
  \centering
  \includegraphics[width=\linewidth]{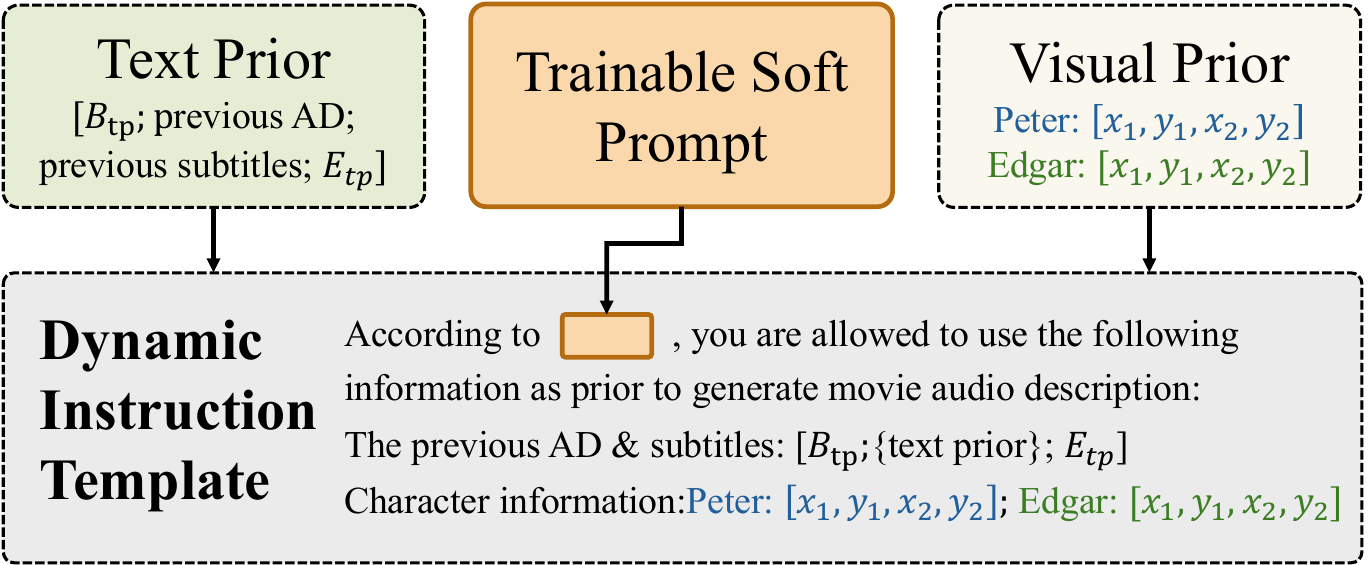}
  \caption{Instruction template with soft prompt. We use a well-designed instruction template with trainable soft prompts to inject the text prior and visual prior into Focused Caption Module.}
  \label{fig:prior}
\end{figure}

In~\cref{fig:prior}, we present our Dynamic instruction template along with the soft prompt (a trainable vector) designed for scenes with varying numbers of characters. Our soft prompting method for instructions exhibits several distinctive properties. Unlike standard prompt tuning methods that serve specific tasks, it is designed to adaptively select valuable information from text prior. Our soft prompting approach guides the model toward AD-specific descriptive styles while incorporating contextual information in a more flexible manner, effectively modeling the relationships between characters, scenes, and inter-character dynamics, including differential weighting of character descriptions based on their narrative significance.

\subsection{Focused Caption Module}
\label{sec3.3}
For the input movie clips, we extract a sequence of key frames $x_i=\{\mathcal{I}_1,\mathcal{I}_2,...,\mathcal{I}_K\}$ based on a predetermined frame stride, accompanied by active main character regions. To generate character-level token representations, we use the shared visual encoder to extract frame-level feature $F_{x_i}\in \mathbbm{R}^{K\times H_I\times W_I \times D_I}$ , where $H_I$, $W_I$, $D_I$ denote the height, width and dimension of the image feature and $K$ represents the number of key frames, respectively. Following ~\cite{liu2023llava}, we employ an MLP as a visual connector to map global key frame visual features to interleaved scene visual tokens $S$.

Each binary mask $M$ of a character obtained from  Character Perception Module is then resized to match the shape of the image feature. We utilize the Mask Pooling operation upon image feature to extract character-level spatial feature $F_C \in \mathbbm{R}^{K\times D_I}$ for each mask, which pools all features within the region $M$ to generate a character-level representation. Finally, we employ a MLP layer to adapt and produce the character-level token $C \in \mathbbm{R}^{K\times D_C}$ for each character region.
To aggregate distinct temporal character-level representations across multiple key frames over a time duration while minimizing redundant tokens, we employ the Temporal Token Merge Module~\cite{yuan2024videorefer}, which takes spatial object $C \in \mathbbm{R}^{K\times D_C}$ tokens as input. We first compute the cosine similarity between each pair of adjacent tokens, formulated as:
\begin{equation}
\mathbf{S}_{m,m+1}=\frac{\mathbf{C}_m\cdot\mathbf{C}_{m+1}}{\|\mathbf{C}_m\|\cdot\|\mathbf{C}_{m+1}\|},0\leq m\textless K,
\end{equation}

We predefine a constant $\mu=4$, then select the top $k-\mu$ similarity scores from $S$. Based on the selected similarity scores, we merge the corresponding $k-\mu$ token pairs into $\mu$ unions, where each union contains multiple highly correlated tokens. We perform average pooling on each union to generate $\mu$ representative tokens, which are then processed through MLP layers to produce the final output $C\in \mathbbm{R}^{\mu \times D_C}$.

Finally, the scene-level tokens $\mathcal{T}_S$ , character-level tokens $\mathcal{T}_C$ and text tokens $\mathcal{T}_W$ are sent to the LLM to obtain the current AD $T_i$ , formulated as
$T_i = \Phi(\mathcal{T}_S , \mathcal{T}_C, \mathcal{T}_{W})$, where $\Phi$ denotes the LLM.

\section{Implementation Details}
\subsection{Training Datasets}

\textbf{Storyboard-v2.} Our main objective is to generate audio descriptions for movie clips that focus on character-centric details. 
For this goal, the model need to be trained on the AD datasets. However, existing AD datasets have significant limitations. MAD~\cite{soldan2022mad} only provides video data in the form of CLIP visual features in order to avoid copyright restrictions. The narrations in Movie101~\cite{yue-etal-2023-movie101} lack event-level sentence segmentation, which proves detrimental to the training process. VPCD~\cite{brown2021face} annotates character body trajectories and features, lacking descriptions of character actions. The MCVD~\cite{chen2011collecting} focuses on single, explicit events but lacks contextual information.

\begin{figure}[h]
  \centering
  \includegraphics[width=\linewidth]{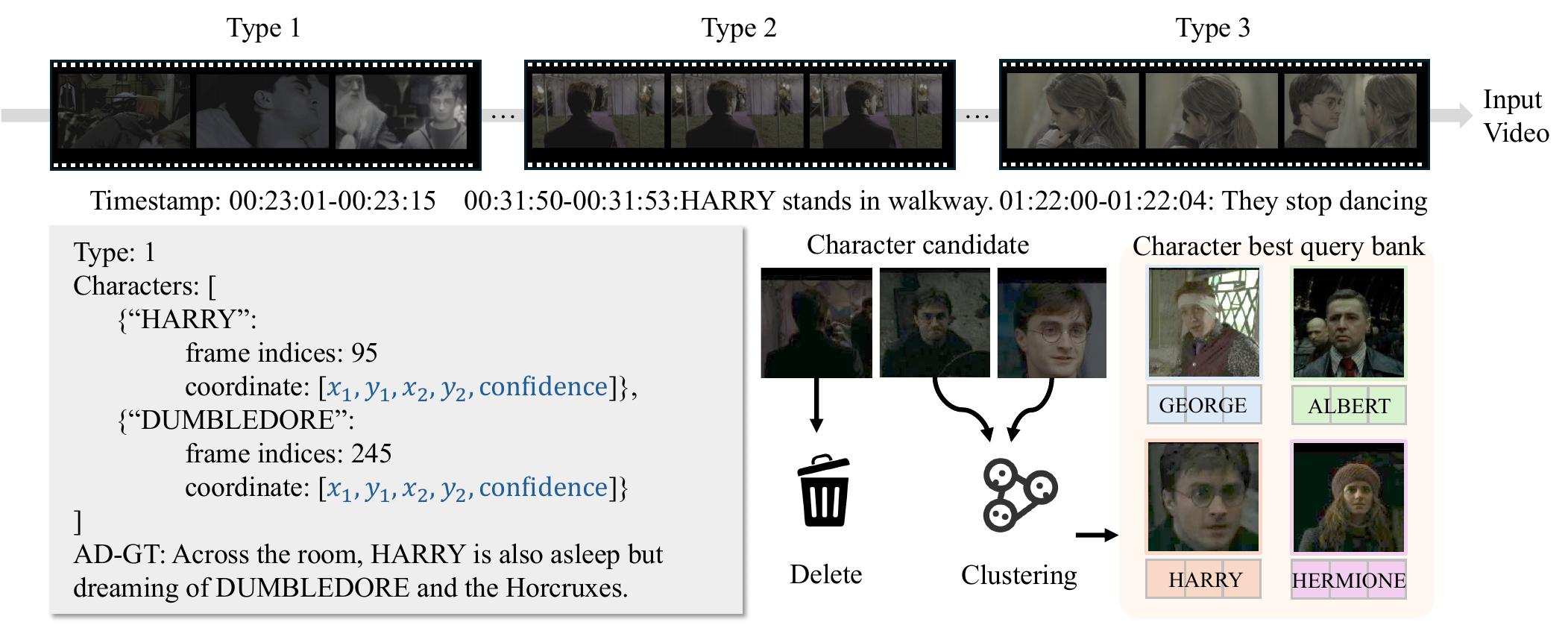}
  \caption{Samples of Storyboard-v2. Our dataset involves three main part, i.e., (i)movie clips, (ii) character regions, (iii) movie audio description ground-truth}
  \label{fig:train data}
\end{figure}

To address these limitations, we design an automated pipeline based on the video generation dataset Storyboard20K~\cite{xie2024learning}, to construct Storyboard-v2 as our training dataset, illustrated in~\cref{fig:train data}. Storyboard20K comprises 150K keyframes sourced from movies in MovieNet~\cite{huang2020movienet} and LSMDC~\cite{rohrbach2017movie}, with rich multi-modal annotations designed to model complex narrative structures in films. Our constructed Storyboard-v2 comprises 11,250 triplets, each containing a no-dialogue movie clip, the corresponding best character query bank for that movie, and movie ad ground truth.

\begin{figure}[h]
  \centering
  \includegraphics[width=1\linewidth]{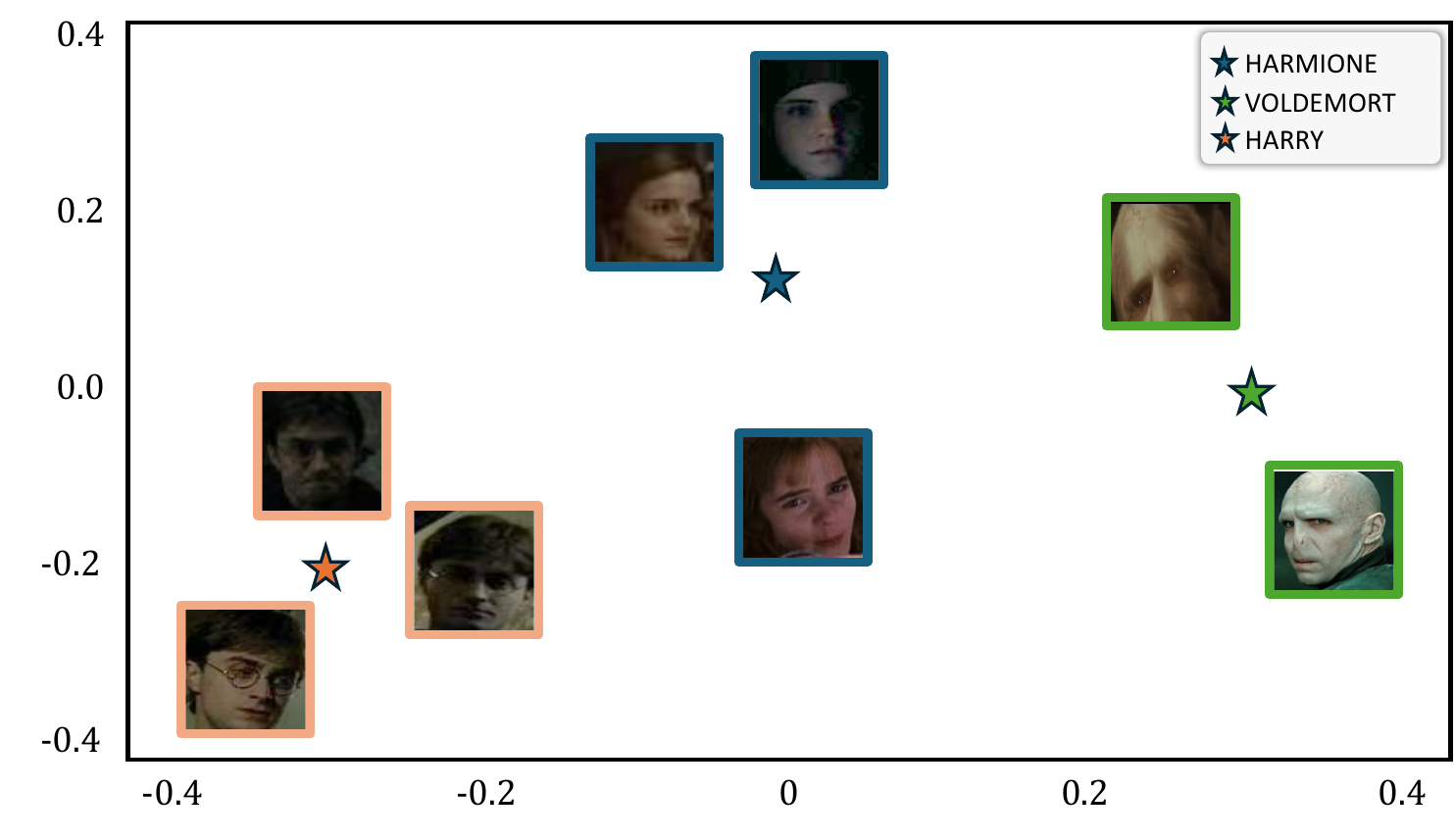}
  \caption{The clustering results for obtaining the best query from the movie \textit{Harry Potter and the Deathly Hallows}.}
  \label{fig:cluster}
\end{figure}

Specifically, our automated AD collection pipeline consists of four stages:
(i) Given that Storyboard20K only provides a single keyframe per movie clip, making it challenging to directly apply to audio description model training, we initially extract complete movie clips from the full-length films based on the timestamps to obtain segments requiring audio description.
(ii) To address the lack of character name-to-feature mapping information in existing datasets and the domain gap between IMDB actor images and character appearances that hinders audio description generation, we utilize Storyboard20K's coarse-grained character annotations. Despite limitations (e.g., some character bounding boxes only encompass partial body regions and suffer from blurring issues), we first crop all character-annotated regions and employ FaceNet~\cite{schroff2015facenet} to filter out non-facial regions. For the remaining facial images, we apply the clustering algorithm described in ~\cref{sec3.1} to compute best queries that serve as character representations. These representative embeddings are then used for character recognition across all movie clips. ~\cref{fig:cluster} demonstrates the clustering results for obtaining the best query from the movie \textit{Up in the Air}. This approach effectively mitigates feature shifts caused by discrepancies between portraits and actual movie appearances (e.g., hairstyle, age, camera angles) and variations in actor appearance across different scenes.
(iii) Additionally, based on the complete movie narrative arc, we collect previous audio descriptions to serve as contextual priors.
(iv) We categorize AD annotations into three distinct types:
\begin{itemize}
    \item Type 1: ADs with matching character names and visual regions, used to train the model's focused region description capabilities;
    \item Type 2: ADs containing character names without corresponding visual features in video frames, used to train the model's contextual feature utilization;
    \item Type 3: ADs without character mentions but containing main characters in keyframes, used to train the model's dynamic character weighting through the soft prompt.
\end{itemize}

\subsection{Test Datasets}

\textbf{MAD-eval-Named}~\cite{han2023autoad}  is constructed from a quality-filtered set of 162 movies in the LSMDC dataset~\cite{rohrbach2017movie}. After excluding all films present in both LSMDC training and test sets, MAD-eval-Named retained 10 completely independent movies as our evaluation subset. This dataset preserves the original character name information provided by LSMDC and utilizes non-anonymized AD annotations, comprising video segments with an average duration of 3-5 seconds along with their corresponding audio descriptions.

\textbf{Cinepile-AD} is derived from the visual description component of the CinePile benchmark~\cite{rawal2024cinepile} . CinePile's visual descriptions are human-generated narrations explaining critical visual elements in movie scenes, originally created for audio accessibility. These descriptions were aligned with video clips via ASR, ensuring temporal and contextual consistency. 
The Cinepile dataset is sourced from YouTube through a systematic collection process. We first obtain source movie clips from YouTube and employ WhisperX~\cite{bain2023whisperx} to transcribe subtitles from the complete audio track. The segments without dialogue are identified as narrative movie clips requiring audio description. We then extract the corresponding visual description annotations as ground truth based on the surrounding subtitle context. Additionally, we collect previous audio descriptions and subtitles as contextual information for each segment.

\textbf{Post-Processing.} The character information for movies can be collected from online databases or review websites like IMDb4. In detail, for each movie in MAD-eval-Named and Cinepile-AD datasets, we download the top 5 to 8 cast information from IMDb including the actor names, their character role name, and the actor portrait image, and implement character query clustering following the approach detailed in ~\cref{sec3.1}. To ensure the model does not rely on acoustic information, we removed all audio content from the movie clips.

Given the memory constraints of LLM, and as reported in ~\cite{han2023autoad} that the trend for the recurrent setting flattens when the context ADs are longer than 3 sentences, we collect three preceding audio descriptions along with interleaved subtitles as text prior for each segment in our dataset.

\subsection{Architecture}
We use a feature pyramid network~\cite{lin2017feature} to fuse the stride 16 and 32 features of the Hiera image encoder respectively to produce the image embeddings for each frame and character regions. 
\textbf{Character Perception Module}(~\cref{sec3.1}) takes the movie character features $\{q_i\}$ and the movie clip as input, and outputs the names of active characters in movie clips, their corresponding keyframes of appearance, and associated region coordinates. We employ FaceNet~\cite{schroff2015facenet} for face verification and recognition with a memory attention mechanism and a memory encoder-bank system. The memory attention utilizes both positional and rotary embeddings across L=4 layers. The memory system optimizes computational efficiency by reusing Hiera encoder embeddings and maintaining a structured memory bank for cross-attention operations.
For \textbf{Focused Caption Module}(~\cref{sec3.3}), we adopt Qwen2-7B-Instruct~\cite{qwen2} as the LLM.

\subsection{Training and Inference Details}
On the Storyboard-v2 datasets, we use a batch size of 8 sequences, each of which contains 16 consecutive character region-movie clips-AD triples from a movie. Overall that gives $8 \times 16$ triples for every batch. From each video clip, we employ a frame stride of 15 to extract keyframes. 

The AdamW~\cite{loshchilov2016sgdr} is used as the optimizer and the cosine annealing scheduler~\cite{loshchilov2017decoupled} is used to adjust learning rate. The learning rate is $2\times 10^{-5}$ with a global batch size of 128 for one epoch. We conduct training using 4 NVIDIA H20 GPUs.
For text generation, greedy search and beam search are commonly used sampling methods. We stop the text generation when a full stop mark is predicted, otherwise we limit the sequence length to 60 tokens. We use beam search with a beam size of 5 and mainly report results by the top-1 beam-searched outputs.
\section{Experiments}

The experimental section is organised as follows: we start by describing the evaluation metrics in ~\cref{sec:5.1}; then in ~\cref{sec:5.2}, we demonstrate the effectiveness of our proposed architecture based on
the groundtruth AD time segments from test datasets, for example, effect of Character Perception Module, Soft prompt and Focused Caption Module; in ~\cref{sec:5.3}, we present qualitative results, and compare our FocusedAD with previous SOTAs in ~\cref{sec:5.3}.

\subsection{Evaluation Metrics}
\label{sec:5.1}
\textbf{Classic Metrics.} To evaluate the quality of text compared with the ground-truth, we use existing tool~\cite{sharma2017nlgeval} to calculate classic captioning metrics to compare the generated AD to the ground-truth AD, namely SPICE~\cite{anderson2016spice} (S) ,METEOR~\cite{banerjee2005meteor} (M) and BertScore~\cite{zhang2019bertscore} (BertS). 

\textbf{Redundancy-aware Metric.} We propose a novel semantics-based metric (R) to evaluate redundancy in generated descriptions. Given a ground-truth annotation sequence $\mathbf{A} = [w^a_1, w^a_2, \dots, w^a_m]$ and a model-generated description $\mathbf{B} = [w^b_1, w^b_2, \dots, w^b_n]$ for the same video clip (where $m$ and $n$ denote the respective lengths after encoding), our metric identifies redundant text segments in $\mathbf{B}$ by comparing their semantic similarity to $\mathbf{A}$. Specifically, if a word vector in $\mathbf{B}$ exhibits a similarity score below a predefined threshold $\theta$ with all vectors in $\mathbf{A}$, it is flagged as redundant. To ensure robustness, our word vectors are derived from a rigorously validated vocabulary, excluding discourse markers and auxiliary words. This approach quantifies how concisely a model captures GT semantics. The redundancy score is computed as follows:

\begin{align}
\text{R}(w_b \rightarrow A) &= 
\begin{cases}
1 - \max\limits_{w_a \in A} \mathrm{sim}(w_a, w_b), & \text{if } \max \mathrm{sim}(w_a, w_b) < \theta \\
0, & \text{otherwise}
\end{cases}
\label{eq:redundancy_word} \\
\text{R}(B \rightarrow A) &= \frac{1}{|B_{\text{valid}}|} \sum_{w_b \in B_{\text{valid}}} \text{R}(w_b \rightarrow A)
\label{eq:redundancy_sentence}
\end{align}

,where $w_a$ and $w_b$ denote word/token vectors from $\mathbf{A}$ and $\mathbf{B}$, respectively; $\theta$ is the similarity threshold; $|B_{\text{valid}}|$ represents the number of valid (content-bearing) words in $\mathbf{B}$.

\subsection{Audio Description on GT Movie Clips}
\label{sec:5.2}
This section focuses on the effectiveness of each proposed component in the FocesedAD pipeline, based on the ground-truth AD time movie clips, as shown in ~\cref{tab:ablation}.

\begin{table}[htbp]
\centering
\small 
\begin{tabular}{cccccccc}
\toprule
Exp. & TP & character bank & SP & Region & S & M & BertS \\
\midrule
A1 & \ding{51} & best query & \ding{55} & \ding{55} & 5.2 & 7.6 & 54.7 \\
A2 & \ding{51} & best query & \ding{51} & \ding{55} & 6.2 & 7.6 & 55.4 \\
\midrule
B1 & \ding{51} & none & \ding{51} & \ding{55} & 2.6 & 6.7 & 54.2 \\
B2 & \ding{51} & actor & \ding{51} & \ding{51} & 4.8 & 7.2 & 54.6 \\
\midrule
C1 & \ding{55} & best query & \ding{51} & \ding{51} & 6.4 & 7.9 & 55.9 \\
C2 & \ding{51} & best query & \ding{51} & \ding{51} & 7.4 & 8.1 & 57.7 \\
\bottomrule
\end{tabular}
\caption{Ablation studies for AD generation on the MAD-eval-Named. Performance is reported in terms of SPICE (S), METEOR (M) and BertScore (BertS). SP means soft prompt. TP means text prior.}
\label{tab:ablation}
\end{table}

\textbf{Effect of Character Perception Module.}
By default, the model takes the predicted active characters in the scene
from the Character Perception Module. From the comparison of rows `B1', `B2' and `C2', we can draw the following observations: (i) injecting character names gives a clear performance gain (`C2 vs B1'), highlighting the dependency of
the AD task on character names; (ii) employing K-means clustering to generate the best query list for characters significantly enhances the accuracy of character recognition and consequently improves the quality of movie audio descriptions (`C2 vs B2'). This comparison also shows the necessity and effectiveness of our Character Perception Module.

\begin{figure}[h]
  \centering
  \includegraphics[width=\linewidth]{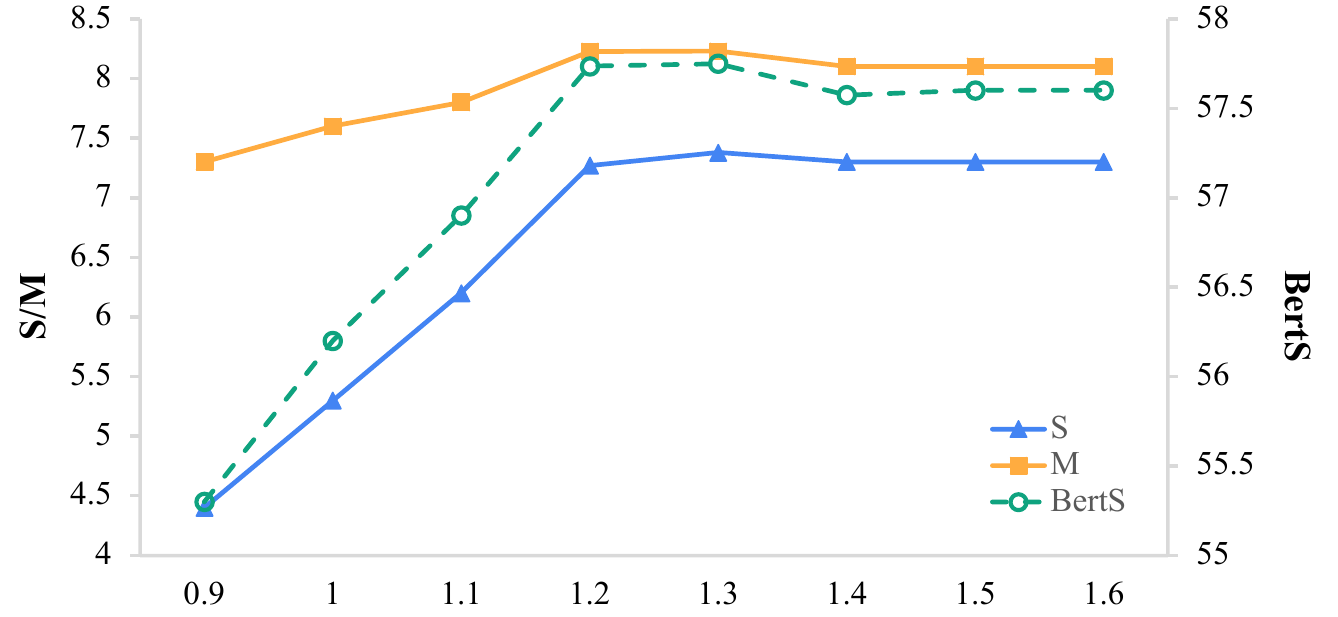}
  \caption{Ablation study on the film \textit{Les Misérables} to evaluate changes in FocusedAD indicators under varying thresholds. This film is selected for its representative nature, as its metrics closely align with the average of MAD-eval-Named.}
  \label{fig:threshold}
\end{figure}

In ~\cref{fig:threshold}, we demonstrate the impact of varying thresholds in Character Recognition on character identification accuracy. We use the model
`C2' in ~\cref{tab:ablation}, and evaluated five candidate thresholds with increments of 0.1. Our experiments reveal that the model achieves optimal character recognition accuracy at a threshold value of $u=1.3$. This optimal threshold likely reflects the common occurrence of profile faces in movies. Setting an excessively high threshold would misclassify main characters as ``other people'', while a threshold that is too low could lead to confusion between different main characters or misidentification of background characters as main characters.

\begin{figure*}[h]
  \centering
  \includegraphics[width=\linewidth]{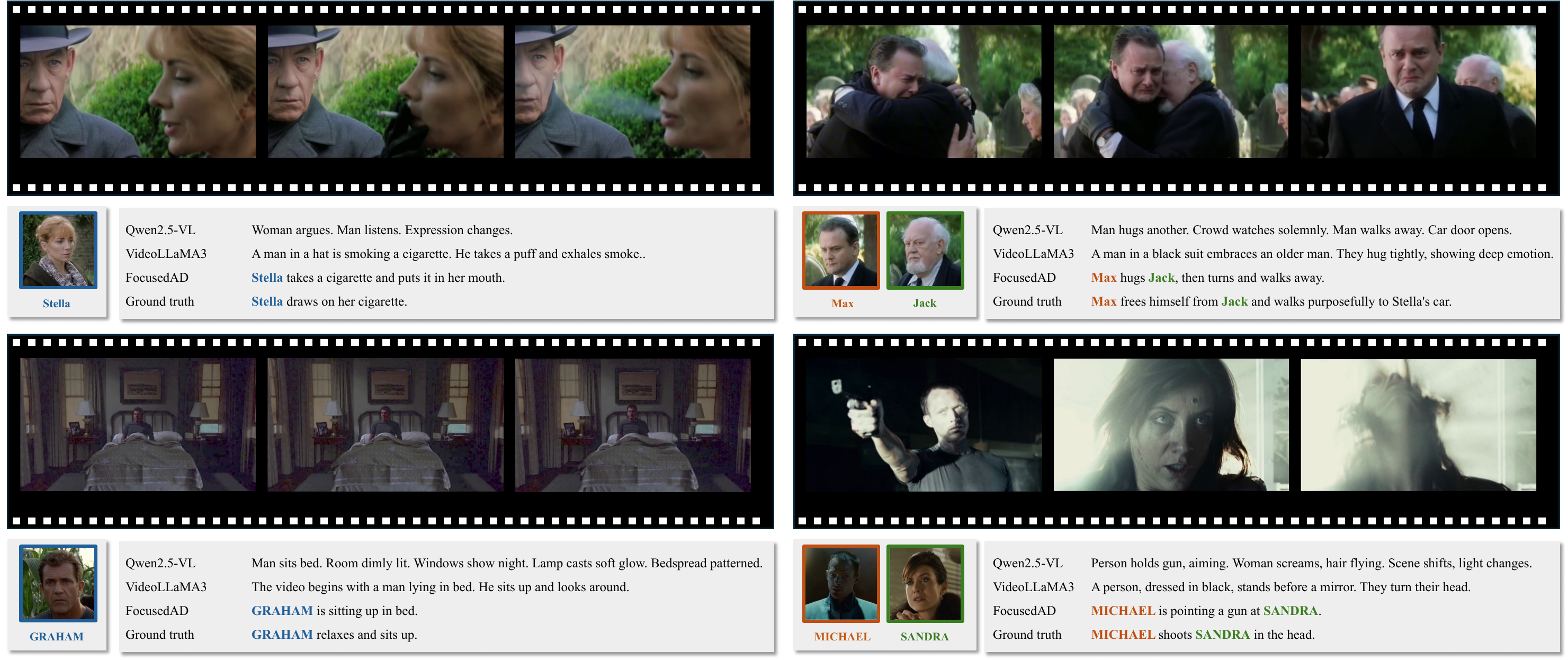}
  \caption{Qualitative results of our method. The top two movie clips demos are from MAD-eval-Named and the bottom two movie clips demos are from Cinepile-AD. The Character Perception Module can recognize active main characters and feed their names into the AD generation pipeline. For visualization purposes, we display the portrait images of characters that have the closest distance to the best query, but the model actually utilizes the best query features as input.}
  \label{fig:qualitative}
\end{figure*}

\textbf{Effect of Dynamic Prior Module.}
In the Dynamic Prior Module, we incorporate previous ADs and subtitles as context for the model (`C1 vs C2'), while introducing a soft prompt mechanism that embeds both text and visual priors into the dynamic instruction template (`A1 vs A2'). The results show that AD generation can be further improved by combining these methods, demonstrating that our Dynamic Prior Module helps the model integrate previous contextual priors to flexibly focus on the content that needs to be described in the current segment when generating ADs.

\textbf{Effect of Focused Caption Module.}
~\cref{tab:ablation} demonstrates the benefits of incorporating the Focused Caption Module (`A2 vs C2'). Our results indicate that generating rich character-level tokens through the object encoder and integrating them with scene-level visual representations and language embeddings effectively guides the model to allocate more attention to main character regions. This mechanism enables the model to focus predominantly on describing main characters' appearances and actions, facilitating the generation of ADs that better focus on narrative-relevant regions.

\subsection{Qualitative Results}
\label{sec:5.3}

~\cref{fig:qualitative} shows four qualitative examples. It shows that Character Perception Module is able to recognize active characters reasonably well, and Focused Caption Module can associate the active characters with the movie audio descriptions.
Given that character names appear in most AD ground truth, with 31\% of movie clips showing character faces at frame 0 and 26\% showing faces in the middle frames, the ability to identify main characters at any frame and propagate this information throughout the movie clips is an essential capability for high-quality AD generation.

\subsection{Comparison with SOTA Approaches}
\label{sec:5.4}
In ~\cref{tab:comparison sota}, we compare our method with previous audio description methods  on the MAD-eval-Named benchmark and Cinepile-AD and achieve state-of-the-art performance on both test datasets. We first evaluate the zero-shot capabilities of our FocusedAD on the test set (as our model is fine-tuned on video generation datasets rather than the MAD~\cite{soldan2022mad,han2023autoad} dataset) against the fine-tuning-based state-of-the-art models, including AutoAD-I~\cite{han2023autoad} and AutoAD-II~\cite{han2023autoadii}. Moreover, our model, based on a 7B parameter LLM, outperforms MM-Narrator~\cite{zhang2024mm}, which relies on the closed-source GPT-4~\cite{achiam2023gpt} model and extensive in-context learning datasets.

We next compare our model with state-of-the-art general video understanding MLLMs, including Qwen2.5-VL~\cite{bai2025qwen2} and VideoLLaMA3~\cite{damonlpsg2025videollama3}. Despite their supervised fine-tuning on extensive video-relevant fine-grained datasets, these models primarily focus on scene-level features. In contrast, our FocusedAD emphasizes character-centric details, generating predictions that better align with movie audio description preferences.
\begin{table}[htbp]
    \centering
    \small 
    \begin{tabular}{c|cccc|cccc}
        \toprule
        \multirow{2}{*}{Models} & \multicolumn{4}{c|}{MAD-eval-Named} & \multicolumn{4}{c}{Cinepile-AD} \\
        \cmidrule(lr){2-5} \cmidrule(lr){6-9}
        & S & M & BertS & R & S & M & BertS & R\\
        \midrule
        AutoAD-I &  4.4 & 7.4 & 24.2 & - & - & - & - & - \\
        MM-Narrator &  5.2 & 6.7 & - & - & - & - & - & - \\
        Qwen2.5-VL &  3.2 & 4.9 & 50.5 & 55.6 &  3.5 & 5.3 & 51.2 & 53.0\\
        VideoLLaMA3 &  4.1 & 7.5 & 54.6 & 45.2 &  2.9 & 5.9 & 52.9 & 42.4\\
        \textbf{FocusedAD} &  \textbf{7.4} & \textbf{8.1} & \textbf{57.7} & \textbf{44.1} &  \textbf{15.4} & \textbf{14.7} & \textbf{64.5} & \textbf{34.8}\\
        \bottomrule
    \end{tabular}
    \caption{Compared with other works on movie AD generation task on MAD-eval-Named and Cinepile-AD. R denotes redundancy and R is better when lower.}
    \label{tab:comparison sota}
\end{table}

\section{Conclusion}
This paper introduces FocusedAD, a novel framework for automatic movie audio description generation that addresses the critical challenges of character-centric narration and narrative-relevant detail selection. We propose a comprehensive system that integrates character perception, dynamic prior information, and focused caption generation. Our approach propagates character regions across video sequences and generating descriptions that focus on story-critical elements. Additionally, we introduce an automated character query bank construction pipeline that resolves the character identification challenges present in existing AD datasets.
The experimental results demonstrate that FocusedAD achieves state-of-the-art performance across multiple benchmarks, including impressive zero-shot results on MAD-eval-Named and our newly introduced Cinepile-AD dataset. 
This work represents a significant step toward fully automated, high-quality movie audio description generation, bringing us closer to human-level performance in this challenging task.


\bibliographystyle{ACM-Reference-Format}
\bibliography{sample-base}










\end{document}